# ARANet: Attention-based Residual Adversarial Network with Deep Supervision for Radiotherapy Dose Prediction of Cervical Cancer

Lu Wen
School of Computer Science
Sichuan University
Chengdu, Sichuan
wenlu0416@163.com

Wenxia Yin
School of Computer Science
Sichuan University
Chengdu, Sichuan
Yingwenxia@163.com

Zhenghao Feng
School of Computer Science
Sichuan University
Chengdu, Sichuan
fzh_scu@163.com

Xi Wu
School of Computer Science
Chengdu University of Information Technology
Chengdu, Sichuan
wuxi@cuit.edu.cn

Deng Xiong
School of Computer Science
Stevens Institute of Technology
Hoboken, NJ
dxiong@stevens.edu

Yan Wang*
*Corresponding author
School of Computer Science
Sichuan University
Chengdu, Sichuan
wangyanscu@hotmail.com

*Abstract*—Radiation therapy is the mainstay treatment for cervical cancer, and its ultimate goal is to ensure the planning target volume (PTV) reaches the prescribed dose while reducing dose deposition of organs-at-risk (OARs) as much as possible. To achieve these clinical requirements, the medical physicist needs to manually tweak the radiotherapy plan repeatedly in a trial-and-error manner until finding the optimal one in the clinic. However, such trial-and-error processes are quite time-consuming, and the quality of plans highly depends on the experience of the medical physicist. In this paper, we propose an end-to-end <u>A</u>ttention-based <u>R</u>esidual <u>A</u>dversarial <u>Net</u>work with deep supervision, namely ARANet, to automatically predict the 3D dose distribution of cervical cancer. Specifically, given the computer tomography (CT) images and their corresponding segmentation masks of PTV and OARs, ARANet employs a prediction network to generate the dose maps. We also utilize a multi-scale residual attention module and deep supervision mechanism to enforce the prediction network to extract more valuable dose features while suppressing irrelevant information. Our proposed method is validated on an in-house dataset including 54 cervical cancer patients, and experimental results have demonstrated its obvious superiority compared to other state-of-the-art methods.

*Keywords—Radiation Therapy, Dose Prediction, Residual Attention Network, Deep Supervision*

## I. INTRODUCTION

As one of the main methods of cancer treatment, radiotherapy aims to maximize the treatment-gain ratio. This means ensuring that the prescribed tumor volume (PTV) receives the specified dose of radiation with adequate coverage while minimizing unnecessary radiation exposure to the organs-at-risk (OARs). With the development of intensity modulated radiation therapy (IMRT) and volume modulated arc therapy (VMAT), the overall quality of the radiotherapy plan has been improved rapidly[1], [2]. However, designing a clinically acceptable treatment plan is still a complicated and extremely time-consuming process, which is highly dependent on the experience of medical physicists [3], [4]. One solution for this limitation is to automatically predict the dose distribution of radiotherapy by extracting knowledge from existing plans, thereby providing the standard for dosimetry verification and quality control of future radiotherapy plans [5-7].

Existing dose distribution prediction methods of radiotherapy planning can be roughly divided into two categories: knowledge-based planning (KBP) [8], [9] and deep neural network (DNN)-based planning [10-14] Generally speaking, KBP methods take the treatment plans of historical patients as references and perform dosimetric predictions for new patients. To be specific, these methods first build a mathematical model for the historical dose information of existing patients and then apply the generated model to predict the dose distribution for new patients. However, these KBP methods highly rely on the handcrafted features corresponding to the dose distribution results, i.e., overlapping volume histogram (OVH), distance-to-target histogram (DTH), and dose volume histogram (DVH), which also come from the manual procedures of feature extraction [15-17].

In recent years, inspired by the great success of deep learning [18-22], DNNs have been successfully applied to predict the dose distribution in radiotherapy [23-26]. For instance, Kearney et al. [23] proposed a DoseNet based on U-net to obtain dose maps efficiently. Song et al. [24] designed a flexible framework based on the deep neural network DeepLabv3+, achieving the automatic prediction of dose maps. Nguyen et al. [25] applied the U-net and DenseNet framework to the dose prediction task and realized the radiotherapy dose planning of the head and neck (H&N) cancer automatically. Compared with the knowledge-based methods, these DNN-based works directly train end-to-end deep models from the historical plans without physicist intervention. Based on the automatically learned features using the deep model, this kind of method can achieve a more effective automatic prediction with relatively high prediction accuracy.

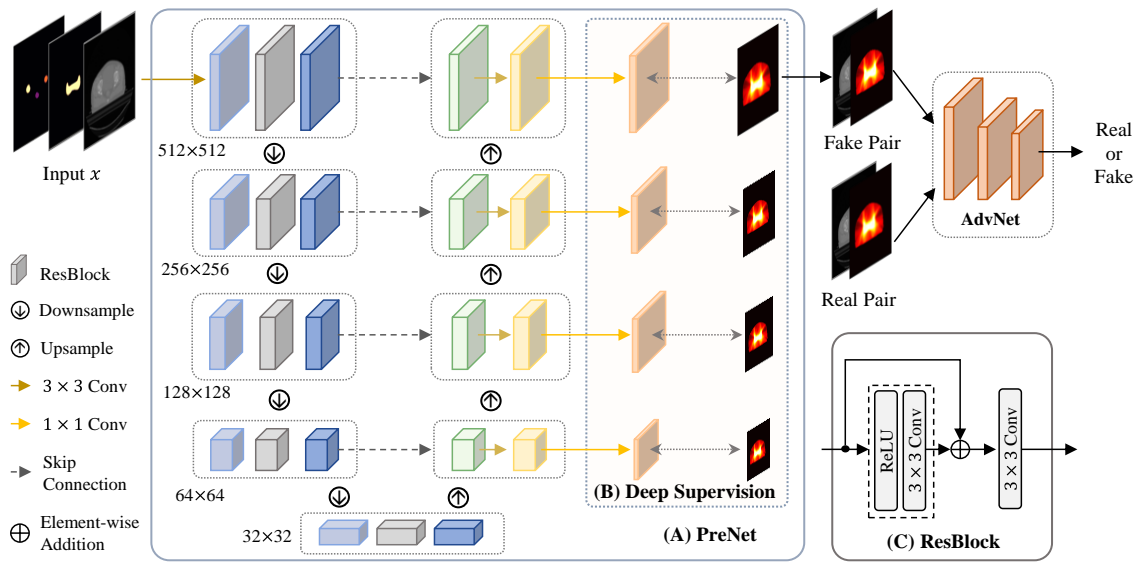

Fig. 1. Illustration of our proposed ARANet which contains a multi-scale residual attention network and a deep supervision mechanism.

In this paper, we develop a 3D dose prediction framework based on an end-to-end residual attention adversarial network with deep supervision, namely ARANet. Fed with CT images and corresponding segmentation masks of PTV and OARs, ARANet can automatically output the dose distribution of cervical cancer. Our contributions can be summarized as:

- Due to the remarkable progress of generative adversarial networks (GANs), the proposed method takes account of an adversarial network (AdvNet), enforcing the prediction network (PreNet) to generate dose maps similar to the real ones.
- To adaptively integrate the valuable dose features while suppressing irrelevant information, a multi-scale residual attention network combined with a deep supervision mechanism is proposed as the generator of GANs.
- Both the quantitative and qualitative experimental results show that our method outperforms the state-of-the-art dose prediction methods.

## II. RELATED WORKS

### A. Knowledge-Based Planning Methods

Knowledge-based planning has made great progress in radiotherapy plans. It predicts the dose distribution of new patients with a set of existing historical plans of treated patients for reference. Based on the indication of dose volume histogram (DVH), Wu *et al*. [15] proposed a new concept, i.e., overlap volume histogram (OVH) which has a strong correlation with DVH. Under the guidance of clinical experience, it was supposed that the farther the distance from the PTV, the lower dose of voxel. Obeying this hypothesis, the corresponding upper (lower) limit can be found according to the OVH comparison of a certain organ. Consequently, the new patient can obtain the DVH prediction of a certain organ by substituting the DTH to the correlation model. To a certain extent, this method can get a good prediction result. However, such knowledge-based methods are prone to two limitations. The first one is that they ignore the three-dimensional spatial information of medical images since the DVH or DTH is the embodiment of two-dimensional information. The second one is that they require much effort and time from medical physicists to determine the applicable handcrafted features to obtain a good prediction result.

### B. DNN-based Planning Methods

Recent research on 3D dose distribution prediction is mostly based on deep neural networks. Specifically, Nguyen *et al*. [25] proposed a new model named hierarchically densely connected U-net (HD U-net) to predict the dose distribution. Specifically, 3D image patches were fed into a U-net framework that integrates dense connecting blocks achieving satisfactory experimental results. Kearney *et al*. [23] presented a prediction model called DoseNet to predict the dose distribution of prostate cancer patients. DoseNet was a fully convolutional neural network based on U-net which accepts six three-dimensional matrices as inputs to the first convolution layer, including three-dimensional CT images of the prostate, bladder, bulbar penis, urethra, and rectum. All input channels were normalized as six separate groups before training. DoseNet finally outputs the prediction results of radiotherapy dose distribution of the patients. Song *et al*. [24] applied the DeepLabv3+ framework to the dose prediction task of rectal cancer. In this work, an atrous spatial pyramid pooling module is used to extract features, and the loss function is defined by mean square error. To address the gradient vanishing problem, Fan *et al*. [26] proposed a dose prediction scheme based on ResNet, and demonstrated that deep learning methods are beneficial to clinical applications, as the quality and efficiency of treatment planning for radiotherapy are improved.

## III. METHODOLOGY

### A. Architecture

The overview of our framework is illustrated in Fig.1 which considers the generative adversarial network as the backbone, including a generator (PreNet) and a discriminator (AdvNet). To extract valuable dose features while suppressing irrelevant information, a multi-scale residual attention module is designed in the PreNet. Also, to speed up the convergence process and improve prediction accuracy, a deep supervision mechanism is integrated with PreNet. On the whole, in the training stage, the original CT images, segmentation images of OARs (including bladder, left and right femur, small intestine), and PTV are fed into the neural network. With the help of the competition between the AdvNet and the PreNet, we can finally get a trained-well model that can output a predicted 3D dose distribution image of cervical cancer patients. In the test stage, we only need the trained model to predict new cases and discard the AdvNet. Please note that, the clinically acceptable dose distribution images which are marked by the physicists are regarded as the ground truth for network training.

*B. PreNet*

For the PreNet, we first concatenate the slices of 3D CT, OARs, and PTV with the size of 512×512 and input them into the encoding path of the U-net-like network. After a 3×3 kernel with a stride of 1, a 16×512×512 feature map is obtained, which will be further processed by a residual convolution module (shown in Fig. 1(c)) to get a refined result. The refined feature representation is integrated by both high- and low-level features and has the same size as the previous one. Then we take the down-sampling operation to decrease the scale of the feature map while increasing the number of channels by a factor of 2. After the four above-mentioned operations, we can finally obtain an encoded deep feature with the size of 64×64. To map the encoded feature to the dose map space, it will be fed into a decoding path. Going through four upsampling operations, the encoded deep feature can be eventually restored to the same size as input images.

*C. Spatial Attention Mechanism*

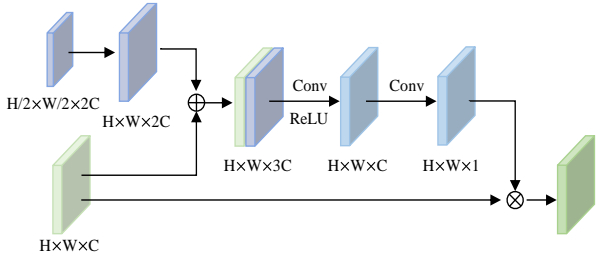

**Fig. 2.** Illustration of spatial attention mechanism inside the skip connections.

As shown in Fig. 2, we use skip connections with a spatial attention mechanism to fuse the lower-layer encoding features with the corresponding higher-layer decoding features during each upsampling stage. Specifically, the decoding deep feature is upsampled to a certain scale that is consistent with the encoding feature, and then cascaded with the encoding feature along the channel direction. After two consecutive convolution layers, the cascaded feature is turned into an attention map, which is leveraged to perform element-wise multiplication with the encoding feature in each channel, thus producing the upper-layer decoding feature. Furthermore, to get a more precise result, we apply 1×1 kernels to acquire coarse predictions at each scale and enforce them to be supervised by the real dose map rescaled to the corresponding size, which is called a deep supervision mechanism.

*D. Loss Functions*

The total loss function $L_{total}$ for network optimization is comprised of two parts: (1) $L_{pre}$ for the PreNet training, and (2) $L_{adv}$ for the AdvNet training. $L_{total}$ can be written as below:

$$L_{total} = \lambda_1 L_G + \lambda_2 L_{adv}, \quad (1)$$

where $\lambda_1$ and $\lambda_2$ are two hyper-parameters to balance the two terms.

Concretely, $L_{pre}$ in Equation (2) contains the deep supervision loss $L_{DS}$ and the loss $L_{final}$ between the final predicted dose map and the ground truth.

$$L_G = L_{final} + \lambda_3 L_{DS}, \quad (2)$$

where $\lambda_3$ is a weighted hyper-parameters

As formulated in Equation (3), $L_{DS}$ is a commonly used mean squared error (MSE) loss that supervises the deep predictions at three scales to resemble the ground truth:

$$L_{DS}(y_i, y_i^p) = \sum_{i=1}^{3}(y_i - y_i^p)^2, \quad (3)$$

where $y_i^p$ represents the predicted dose distribution map at i-th supervised scale and $y_i$ is the corresponding ground truth at i-th scale.

$L_{final}$ is a smooth L1 loss which can be expressed as Equation (4), ensuring that the final predicted dose map $G(x)$ stay as close to the real dose map $y$ as possible:

$$L_{final}(y, f(x)) = \begin{cases} \frac{1}{2}(y - G(x))^2, & |y - G(x)| < \delta \\ \delta(|y - G(x)| - \frac{1}{2}\delta), & |y - G(x)| \geq \delta \end{cases}, \quad (4)$$

where $\delta$ represents a prediction error.

Meanwhile, to obtain a more precise result, the AdvNet $D$ takes the adversarial loss for the objective function, which is defined in Equation (5):

$$L_{adv}(D, G) = E(D(y)) + E\left(1 - D(G(x))\right). \quad (5)$$

IV. EXPERIMENTS

*A. Dataset*

To validate the performance of the proposed network, we use an in-house clinical cervical cancer dataset with 54 cases who have taken the VAMT plan in West China Hospital, Sichuan University, Chengdu, China. For each case, the CT scan and corresponding segmentation mask of OARs are included. Specifically, the OARs of cervical cancer contain bladder, left and right femur, small intestine, and rectum. For data partition, we randomly select 40, 6, and 8 cases for model training, validation, and testing, respectively.

*B. Dosimetric evaluation metrics*

To objectively evaluate the results of the proposed dose prediction model, we calculate the normalized average percent prediction error (APE) to measure the difference between the metrics of ground truth and prediction. The mathematical definition is shown in Equation (6).

$$\Delta = \frac{1}{n}\sum_{i=1}^{n} \frac{|Truth_i - Prediction_i|}{Prediction_i} \times 100\%. \quad (6)$$

Simultaneously, following [25], we use the conformity index (CI) and heterogeneity index (HI) to evaluate the dosimetric distribution inside the PTV areas between the predicted dose distribution map and the ground truth. Among them, CI is formulated as Equation (7), where $TV$ is the target volume, $PIV$ is the prescription isodose volume.

TABLE I. QUANTITATIVE COMPARISON WITH SOTA METHODS IN TERMS OF PTV $D_{mean}$ AND PTV V50, RESPECTIVELY. THE BEST RESULTS ARE IN **BOLD**.

| | PTV Dmean (Gy) | | | | | PTV $V_{50}$ | | | | |
|---|---|---|---|---|---|---|---|---|---|---|
| *Patient* | *Original* | *Proposed* | *ResNet* | *DoseNet* | *DeepLabv3+* | *Original* | *Proposed* | *ResNet* | *DoseNet* | *DeepLabv3+* |
| P1 | 45.76 | 46.17(0.41) | 51.29(5.53) | 45.98(0.22) | 48.75(3.00) | 0.13 | 0.18(0.04) | 0.94(0.80) | 0.74(0.60) | 0.66(0.52) |
| P2 | 45.33 | 45.80(0.48) | 51.19(5.87) | 45.17(0.15) | 47.94(2.61) | 0.06 | 0.08(0.02) | 0.94(0.89) | 0.27(0.21) | 0.65(0.60) |
| P3 | 45.64 | 45.91(0.27) | 51.96(6.32) | 44.74(0.90) | 49.15(3.51) | 0.05 | 0.08(0.02) | 0.99(0.94) | 0.00(0.05) | 0.58(0.53) |
| P4 | 45.80 | 46.26(0.28) | 51.30(5.32) | 45.44(0.54) | 48.98(3.00) | 0.07 | 0.08(0.01) | 0.96(0.89) | 0.66(0.60) | 0.74(0.67) |
| P5 | 45.41 | 45.63(0.22) | 51.98(6.57) | 45.21(0.20) | 48.90(3.50) | 0.02 | 0.02(0.00) | 0.99(0.97) | 0.00(0.01) | 0.48(0.46) |
| P6 | 53.38 | 53.30(0.09) | 51.97(1.41) | 46.13(7.25) | 48.97(4.41) | 0.98 | 0.98(0.00) | 0.99(0.00) | 0.6(0.39) | 0.74(0.25) |
| P7 | 45.99 | 46.19(0.20) | 51.30(5.31) | 45.55(0.43) | 48.26(2.27) | 0.11 | 0.16(0.05) | 0.93(0.81) | 0.37(0.26) | 0.56(0.45) |
| P8 | 52.72 | 51.16(1.55) | 51.92(0.79) | 44.86(7.86) | 49.65(3.06) | 0.96 | 0.80(0.17) | 0.97(0.00) | 0.43(0.54) | 0.85(0.11) |
| Avg | 48.00 | **47.55(0.03)** | 51.61(4.09) | 45.38(2.13) | 48.82(1.30) | 0.30 | **0.30(0.00)** | 0.96(0.67) | 0.38(0.09) | 0.66(0.34) |

TABLE II. QUANTITATIVE COMPARISON WITH SOTA METHODS IN TERMS OF CONFORMITY INDEX (CI) AND HETEROGENEITY INDEX (HI), RESPECTIVELY.

| | CI | | | | | HI | | | | |
|---|---|---|---|---|---|---|---|---|---|---|
| *Patient* | *Original* | *Proposed* | *ResNet* | *DoseNet* | *DeepLabv3+* | *Original* | *Proposed* | *ResNet* | *DoseNet* | *DeepLabv3+* |
| P1 | 0.72 | 0.78(0.06) | 0.58(0.14) | 0.78(0.06) | 0.71(0.02) | 0.29 | 0.21(0.08) | 0.12(0.17) | 0.13(0.16) | 0.16(0.13) |
| P2 | 0.81 | 0.83(0.02) | 0.53(0.28) | 0.78(0.04) | 0.71(0.10) | 0.25 | 0.18(0.07) | 0.11(0.14) | 0.13(0.12) | 0.15(0.10) |
| P3 | 0.86 | 0.88(0.02) | 0.59(0.27) | 0.64(0.22) | 0.72(0.13) | 0.23 | 0.13(0.11) | 0.08(0.15) | 0.15(0.09) | 0.13(0.10) |
| P4 | 0.86 | 0.86(0.00) | 0.64(0.22) | 0.82(0.04) | 0.73(0.13) | 0.25 | 0.17(0.08) | 0.10(0.14) | 0.11(0.14) | 0.14(0.11) |
| P5 | 0.79 | 0.80(0.01) | 0.57(0.22) | 0.79(0.00) | 0.71(0.09) | 0.15 | 0.16(0.01) | 0.08(0.07) | 0.11(0.04) | 0.13(0.02) |
| P6 | 0.64 | 0.56(0.08) | 0.57(0.06) | 0.82(0.18) | 0.69(0.06) | 0.09 | 0.12(0.03) | 0.08(0.01) | 0.12(0.04) | 0.12(0.03) |
| P7 | 0.73 | 0.69(0.05) | 0.55(0.19) | 0.72(0.01) | 0.69(0.05) | 0.29 | 0.25(0.04) | 0.12(0.17) | 0.16(0.13) | 0.18(0.11) |
| P8 | 0.68 | 0.74(0.05) | 0.68(0.02) | 0.71(0.03) | 0.73(0.05) | 0.10 | 0.12(0.02) | 0.10(0.00) | 0.11(0.01) | 0.12(0.02) |
| Avg | 0.76 | **0.77(0.004)** | 0.59(0.17) | 0.76(0.01) | 0.71(0.05) | 0.20 | **0.17(0.04)** | 0.01(0.11) | 0.13(0.08) | 0.14(0.06) |

TABLE III. QUANTITATIVE COMPARISON WITH SOTA METHODS IN TERMS OF BLADDER $D_{mean}$.

| | Bladder Dmean%(Gy) | | | | | | | | |
|---|---|---|---|---|---|---|---|---|---|
| *Patient* | *Original* | *Proposed* | *ResNet* | *Dosenet* | *Deeplabv3+* | *ΔProposed* | *ΔResNet* | *ΔDosenet* | *ΔDeeplabv3+* |
| P1 | 23.929 | 24.560 | 27.998 | 27.745 | 26.971 | 0.631 | 4.069 | 3.816 | 3.042 |
| P2 | 28.768 | 30.207 | 36.555 | 34.800 | 32.722 | 1.439 | 7.787 | 6.032 | 3.954 |
| P3 | 25.433 | 26.699 | 32.364 | 32.730 | 29.832 | 1.266 | 6.931 | 7,297 | 4.399 |
| P4 | 31.201 | 32.254 | 36.090 | 32.798 | 34.008 | 1.053 | 4.889 | 1.597 | 2.807 |
| P5 | 28.322 | 28.738 | 34.879 | 32.408 | 31.514 | 0.416 | 6.557 | 4.086 | 3.192 |
| P6 | 33.972 | 36.992 | 33.678 | 31.875 | 32.463 | 3.02 | 0.294 | 2.097 | 1.509 |
| P7 | 34.757 | 29.882 | 31.816 | 31.473 | 30.374 | 4.875 | 2.941 | 3.284 | 4.383 |
| P8 | 39.151 | 35.820 | 36.317 | 35.693 | 37.153 | 3.331 | 2.834 | 3.458 | 1.998 |
| Avg | 30.691 | **30.644** | 33.712 | 32.44 | 31.879 | **0.047** | 3.02 | 1.749 | 1.188 |

$$CI = (TV \cap PIV)^2 / (TV \times PIV). \quad (7)$$

HI formalism is defined as Equation (8), where $D_m$ is the minimal absorbed dose covering $m\%$ percentage volume of PTV.

$$HI = (D_2 - D_{98})/D_{50}. \quad (8)$$

Since $Vx$ denotes the percentage volume that receives a dose level of at least $x$, we also utilize $V_{50}$ as another metric for dose evaluation.

### C. Training Details

The proposed method is implemented with the PyTorch framework and all experiments are conducted on a single NVIDIA GeForce RTX 3090 GPU equipped with 24GB of memory. The training phase is executed with a batch size of 16. The whole model is trained for 100 epochs where the learning rate is set as 1e-5. An adaptive moment estimation (Adam) optimizer is exploited to facilitate an efficient convergence. As for hyper-parameter $\lambda_1$, $\lambda_2$, and $\lambda_3$, according to the previous experience, we set them as 2, 1, and 0.5 respectively.

### D. Comparison with State-of-the-art (SOTA) Methods

To investigate the superiority of our proposed ARA-Net, we take several SOTA methods into consideration for comparison: (1) ResNet [26], (2) DoseNet [23], and (3) DeepLabv3+ [24]. The quantitative comparison results are summarized in Table I, Table II, and Table III, respectively. Specifically, Table I shows the $D_{mean}$ values and $V_{50}$ values of PTV generated by various methods, where $P_i$ represents $i$-th patient and the values in parentheses represent the difference between the predicted and ground truth. $V_{50}$ denotes the percentage volume that receives a dose level of at least 50. Table II shows the conformity index (CI) and heterogeneity index (HI) compared with the existing methods.

Table I and Table II show the results of indicators on PTV. Compared with other methods, our ARA-Net achieves the best scores on all clinical metrics. Specifically, as displayed in Table I, the proposed method gains 47.55 Gy and 0.30 in terms of $D_{mean}$ and $V_{50}$, which are highly approximated to those of the ground truth, i.e., 48.00 Gy and 0.30. Furthermore, in Table II, such superior performance can also observed in terms of CI, and HI. To more comprehensively evaluate the performance of the proposed method, we show the prediction results on the bladder in terms of $D_{mean}$ in Table III. As seen, our method also achieves the best prediction, i.e., 30.644 Gy, with the least error, i.e., 0.047 Gy.

TABLE IV. ABLATION EXPERIMENT RESULTS IN TERMS OF PTV $D_{mean}$.

| PTV Dmean (Gy) | | | | | | | | | |
|---|---|---|---|---|---|---|---|---|---|
| *Patient* | *Original* | *Proposed* | *U-net* | *AU-net* | *RAU-net* | *ΔProposed* | *ΔU-Net* | *ΔAU-Net* | *ΔRAU-Net* |
| P1 | 45.76 | 46.17 | 50.77 | 46.15 | 47.25 | 0.41 | 5.02 | 0.39 | 1.50 |
| P2 | 45.32 | 45.80 | 49.69 | 43.57 | 46.13 | 0.48 | 4.36 | 1.76 | 0.80 |
| P3 | 45.64 | 45.91 | 49.80 | 45.96 | 45.64 | 0.27 | 4.16 | 0.33 | 0.00 |
| P4 | 45.98 | 46.26 | 49.24 | 43.79 | 47.29 | 0.28 | 3.26 | 2.19 | 1.31 |
| P5 | 45.41 | 45.63 | 49.59 | 44.39 | 47.32 | 0.22 | 4.19 | 1.02 | 1.91 |
| P6 | 53.38 | 53.30 | 50.91 | 46.36 | 47.68 | 0.09 | 2.48 | 7.03 | 5.70 |
| P7 | 45.99 | 46.19 | 49.87 | 44.06 | 45.99 | 0.20 | 3.89 | 1.93 | 0.00 |
| P8 | 52.72 | 51.16 | 50.55 | 45.81 | 48.71 | 1.55 | 2.17 | 6.91 | 4.00 |
| Avg | 47.52 | **47.55** | 50.05 | 45.01 | 47.00 | **0.03** | 2.53 | 2.51 | 0.52 |

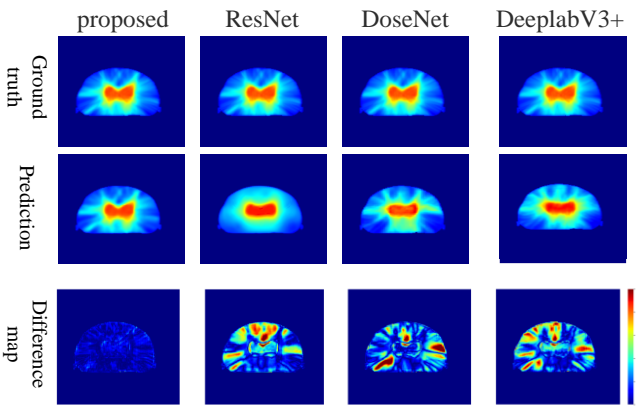

**Fig. 3.** Visualization results compared with SOTA methods. The first row is ground truth, the second and the last rows are dose distribution prediction results and corresponding difference maps respectively.

On the other hand, for qualitative comparison, we provide the dose maps predicted by different methods as well as corresponding difference maps in Fig. 3. It is obvious that the result of our method is the closest to ground truth, especially for the radial pattern and the planning region of dose distribution. The fact that the difference map of our method is darkest also supports the above conclusion.

Based on all these experimental results, we can gain the conclusion that our method achieves superior performance both quantitatively and qualitatively.

*E. Ablation experiments*

To evaluate the effectiveness of every module in ARA-Net. A series of ablation experiments are conducted. The experimental arrangements can be summarized as: (A) U-Net with deep supervision as backbone (denoted as U-Net); (B) U-Net + adversarial training (denotes as AU-Net); and (C) AU-Net + residual attention (denotes as RAU-Net). Corresponding experimental results are exhibited in Fig. 4, Table IV, and Table V. Table IV shows the results of experiments on $D_{mean}$ of PTV. As seen, with the addition of residual attention, the prediction in terms of $D_{mean}$ is enhanced from 45.01 Gy to 47.00 Gy. The average percent prediction error is shown in Table V. Especially in Table V when combining all the designed modules, our method reduces the prediction error of $D_{50}$ to 0±0.007%.

Fig. 4 shows the visualization results of ablation experiments. Specifically, the predicted dose distribution map is extremely similar to the ground truth demonstrating the accuracy of the proposed method.

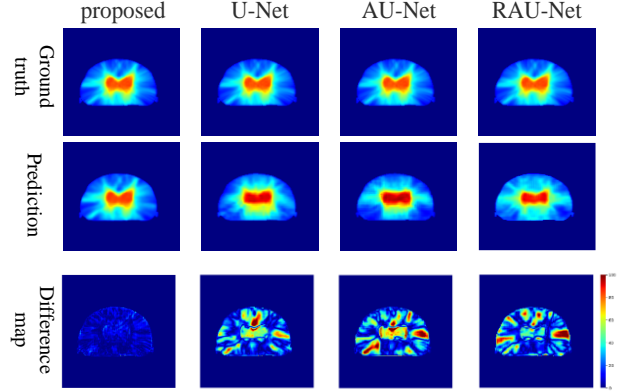

**Fig. 4.** Visualization results of ablation experiments. The first rows are ground truth, the second and last rows are dose distribution prediction results and difference maps respectively.

TABLE V. AVRAGE PERCENT PREDICTION ERROR.

| Method | $D_{98}$ | $D_{50}$ | $D_{mean}$ |
|---|---|---|---|
| U-net | 0.074±0.042 | 0.054±0.057 | 0.045±0.051 |
| AU-net | 0.033±0.050 | 0.03±0.055 | 0.05±0.049 |
| RAU-net | 0.020±0.038 | 0.006±0.058 | 0.01±0.047 |
| Proposed | **0.007±0.033** | **0.000±0.007** | **0.000±0.011** |

In a nutshell, with the gradual addition of key components, the prediction performance is nearly progressively enhanced, verifying the positive impact of our proposed modules.

V. CONCLUSION

In this paper, we have proposed a novel residual attention adversarial network combined with deep supervision namely ARA-Net, which can be used to predict the dose distribution of cervical cancer from CT images automatically. Owning to the residual convolution module that can extract more detailed features and the attention module that can make full use of features in different channels, we can acquire a prediction dose map that is highly consistent with the real dose distribution without human intervention. Besides, by using deep supervision, we can further refine the predicted dose map to achieve a clinically acceptable level. All the conducted experiments prove that our ARA-Net is advanced in automatic dose distribution prediction.

ACKNOWLEDGMENT

This work is supported by National Natural Science Foundation of China (NSFC 62371325, 62071314), Sichuan Science and Technology Program 2023YFG0263, 2023YFG0025, 2023YFG0101.